\documentclass{article}

\usepackage[preprint]{neurips_2026}

\usepackage[T1]{fontenc}
\usepackage{tikz}
\usepackage[pagebackref=true,colorlinks,allcolors=blue]{hyperref}
\usepackage{url}            
\usepackage{booktabs}       
\usepackage{amsfonts}       
\usepackage{nicefrac}       
\usepackage{microtype}      
\usepackage{xcolor}         
\usepackage{times}
\usepackage{epsfig}
\usepackage{graphicx}
\usepackage{amsmath}
\usepackage{amssymb}
\usepackage{amsthm}
\usepackage{paralist}
\usepackage{pifont}
\usepackage{multirow}
\usepackage{bm}
\usepackage{bbm}
\usepackage{diagbox}
\usepackage{makecell}
\usepackage{float}
\usepackage{stfloats}
\usepackage{algorithm}  
\usepackage{color, colortbl}
\usepackage{arydshln}
\usepackage{marvosym}
\usetikzlibrary{arrows.meta,calc,fit,positioning}
\usepackage{algpseudocode}

\theoremstyle{definition}

\theoremstyle{plain}

\renewcommand{\eqref}[1]{Eq.~(\ref{#1})}

\title{HiDrive: A Closed-Loop Benchmark for High-Level Autonomous Driving}

\author{%
Zhongyu Xia\thanks{These authors contributed equally.  \textsuperscript{\Letter} Corresponding author.} \quad Guanyu Zhu$^*$\quad Guo Tang\quad Wenhao Chen\quad Yongtao Wang\textsuperscript{\Letter}
\\
Wangxuan Institute of Computer Technology, Peking University
\\
\texttt{ \{xiazhongyu,confection\_tang,wyt\}@pku.edu.cn }
\\
\texttt{ \{guanyuzhu,wenhaochen\}@stu.pku.edu.cn}
}

\begin{document}

\maketitle
\begin{abstract}

End-to-end autonomous driving has witnessed rapid progress, yet existing benchmarks are increasingly saturated, with state-of-the-art models achieving near-perfect scores on widely used open-loop and closed-loop benchmarks. This saturation does not mean that the problem has been solved; instead, it reveals that current benchmarks remain limited in scenario diversity, object variety, and the breadth of driving capabilities they evaluate. In particular, they lack sufficient long-tail scenarios involving rare but safety-critical objects and fail to assess advanced decision-making such as legal compliance, ethical reasoning, and emergency response. To address these gaps, we propose HiDrive, a new closed-loop benchmark for end-to-end autonomous driving that emphasizes long-tail scenarios and a richer evaluation of driving capabilities. HiDrive introduces a diverse set of rare objects and uncommon traffic situations, and expands evaluation from basic driving skills to more advanced capabilities, including rule compliance, moral reasoning, and context-dependent emergency maneuvers. Correspondingly, we extend previous collision-avoidance-centered metrics into a comprehensive evaluation system that encompasses collision and braking, traffic-rule compliance, and moral-reasoning indicators. Built on a more advanced physics engine, HiDrive provides physically realistic lighting and high-fidelity visual rendering, offering a more challenging and realistic testbed for assessing whether autonomous driving systems can handle the complexity of real-world deployment. The HiDrive software, source code, digital assets, and documentation are available at 
\url{https://github.com/VDIGPKU/HiDrive}.

\end{abstract}

\section{Introduction}
\label{sec:intro}

Autonomous driving, which aims to transform raw sensor observations into safe and reliable driving decisions, has made remarkable progress in recent years and attracted increasing attention from both academia and industry. 
To measure such progress, autonomous driving benchmarks have gradually evolved from open-loop evaluation, which typically relies on pre-recorded real-world driving data and compares model outputs with human demonstrations, to closed-loop evaluation, which is usually conducted in simulation environments where the ego vehicle’s actions can influence subsequent states and interactions.
Compared with open-loop evaluation, closed-loop evaluation better reflects real-world deployment, because autonomous driving is inherently interactive and sequential. 
Additionally, in a closed-loop simulator, it is possible to construct various long-tail scenarios that are difficult to collect from real-world data.

However, the rapid improvement of recent models has led to clear saturation on these benchmarks, where state-of-the-art methods achieve increasingly high scores, leaving little room for meaningful differentiation. 
For example, on the most widely used open-loop benchmarks, including nuScenes~\cite{nuScenes}, NAVSIM-v1~\cite{navsim}, and NAVSIM-v2~\cite{navsim,cao2025pseudo}, state-of-the-art methods have achieved a collision rate of 0.03\%~\cite{xia2025knowval}, 94.8\% Best-of-N PDMS~\cite{chen2026devil}, and 89.3\% EPDMS~\cite{wang2026latent}, respectively.
On Bench2Drive~\cite{jia2024bench2drive}, one of the most commonly used closed-loop benchmarks, state-of-the-art methods have reached a driving score of 89.2\%~\cite{sun2026sparsedrivev2}.
This saturation does not necessarily indicate that end-to-end autonomous driving has been solved; rather, it reveals that existing benchmarks remain relatively simple in terms of scenario complexity, object diversity, and the range of driving abilities evaluated.

First, current benchmarks include only a limited number of long-tail scenarios, despite their being crucial for safe real-world driving. 
In daily traffic environments, autonomous vehicles may encounter rare but safety-critical objects, such as various accident warning signs, traffic cones, damaged or abandoned public facilities, or irregularly shaped obstacles.
These objects may appear infrequently in standard datasets, but failing to correctly perceive or reason about them can lead to severe consequences.
Existing benchmarks are often dominated by common categories such as cars, pedestrians, cyclists, and traffic signs, which encourages models to optimize for frequent cases while leaving their robustness to rare objects largely underexplored. 
As a result, high benchmark performance may mask serious weaknesses in handling long-tail risks.

Second, existing benchmarks mainly focus on basic driving abilities, such as lane following, obstacle avoidance, traffic light recognition, and collision-free navigation. 
However, real-world driving requires much richer and more nuanced decision-making. 
For example, when a vehicle passes through a waterlogged road section with pedestrians nearby, it should slow down to avoid splashing water onto them. 
Similarly, when an ambulance or police car approaches from behind while the ego vehicle is waiting at a red light, the vehicle may need to make an emergency maneuver to yield, even if this requires temporarily entering an otherwise prohibited area or crossing the stop line under exceptional circumstances. 
These scenarios go beyond simple rule-following: they require an understanding of traffic laws, social norms, moral principles, and the hierarchy of safety-critical decisions. 
Therefore, a comprehensive autonomous driving benchmark should evaluate not only whether a vehicle can drive safely in common cases, but also whether it can reason appropriately in complex, ambiguous, and norm-sensitive situations.

To address these challenges, we propose HiDrive, a new closed-loop benchmark for end-to-end autonomous driving that emphasizes long-tail scenarios and a richer evaluation of driving capabilities. 
Our benchmark introduces a diverse set of rare objects and uncommon traffic situations that are underrepresented in existing planning benchmarks. 
In addition, it expands the scope of evaluation from basic driving skills to more advanced capabilities, including legal compliance, ethical reasoning, emergency response, and context-dependent decision-making.
Accordingly, we extend the previous collision-avoidance-centered evaluation metrics to a comprehensive evaluation system that covers collision and braking, traffic rule compliance, and moral reasoning indicators.
Additionally, we adopt a more advanced physics engine to support physically plausible dynamics, realistic lighting, and high-fidelity visual rendering.
By doing so, the proposed benchmark provides a more challenging and realistic testbed for assessing whether autonomous driving systems can handle the complexity of real-world deployment.

\section{Related Work}
\label{sec:related}



\subsection{Planning Benchmarks for Autonomous Driving}


Autonomous driving benchmark evaluation is gradually evolving from open-loop datasets to integrated planning benchmarks.
In recent years, research has increasingly focused on evaluating the quality of planning for learning-based end-to-end systems under interactive protocols.
Table~\ref{tab:planning_benchmark_compare} compares representative benchmarks from the perspectives of closed-loop property, ability/scenario scale, testing-route volume, and legal/ethical coverage.

\begin{table}[H]
\centering
\caption{\textbf{Planning benchmark comparison.} We compare benchmarks across closed-loop property, ability/scenario scale, test-route quantity, and legal/ethical coverage.}
\label{tab:planning_benchmark_compare}
\small
\setlength{\tabcolsep}{3.5pt}
\resizebox{\linewidth}{!}{
\begin{tabular}{lcccccc}
\toprule
\textbf{Benchmark} & \textbf{Closed-Loop} & \textbf{Abilities} & \textbf{Scenarios} & \textbf{Test Routes} & \textbf{Legal Coverage} & \textbf{Ethical Coverage} \\
\midrule
nuScenes~\cite{nuScenes} & \ding{55} & -- & 1,000 & -- & Collision-only & \ding{55} \\
NAVSIM (v1/v2)~\cite{navsim,cao2025pseudo} & \ding{55} & -- & -- & -- & Collision-only & \ding{55} \\
CARLA Leaderboard v2.0~\cite{dosovitskiy2017carla} & \ding{51} & 6 & 21 & -- & Collision, Red-light & Yielding \\
Bench2Drive~\cite{jia2024bench2drive} & \ding{51} & 5 & 44 & 220 & Collision, Red-light, Speed-limit & Yielding \\
\midrule
\textbf{HiDrive (Ours)} & \textbf{\ding{51}} & \textbf{30} & \textbf{94} & \textbf{330} & \textbf{More comprehensive} & \textbf{Comprehensive} \\
\bottomrule
\end{tabular}
}
\end{table}

Existing benchmarks still have clear limitations, mainly due to their task definitions and evaluation boundaries.
nuScenes~\cite{nuScenes} is a typical open-loop benchmark. Its strength lies in large-scale annotation and perception/prediction alignment. However, its evaluation is based on offline logs, and policy actions do not feed back to change other traffic participants. As a result, it cannot cover the full closed-loop process of ``decision-environment feedback-redecision.'' For sudden, dynamic situations (e.g., a pedestrian appearing from behind an occlusion), open-loop metrics can serve as a reference but cannot directly replace closed-loop conclusions.

NAVSIM~\cite{navsim,cao2025pseudo} extends log-driven simulation to a larger-scale, unified evaluation layer, offering a practical balance between cost and coverage. However, its ``non-reactive'' setup means background traffic mainly follows predefined trajectories, so tested policies have limited influence on other agents' behaviors. This makes NAVSIM suitable for large-scale screening and iteration, but less suitable for evaluating strong interaction capabilities with other participants (e.g., aggressive cut-ins or continuous yielding).

CARLA Leaderboard v2.0~\cite{dosovitskiy2017carla} and Bench2Drive~\cite{jia2024bench2drive} both provide sensor-level closed-loop evaluation, allowing direct assessment of route completion under infraction penalties. Their main limitations are twofold. First, scores are mostly route-aggregated, ability decomposition remains relatively coarse, and evaluation signals are relatively single, mainly collision-centered safety penalties and traffic-violation-centered responsibility penalties. It is therefore difficult to measure whether a model both avoids over-interfering with surrounding traffic and identifies potential risks early. Second, the simulation world still differs significantly from the real world, including limited asset diversity, constrained rendering realism, and insufficient coverage of long-tail behaviors (e.g., rule-breaking or abnormal traffic participants). Meanwhile, as of May 2026, the official CARLA AD Leaderboard lists v2.1 as the current version, keeps older versions separately supported, and states that the 2025 CARLA AD Challenge did not run~\cite{dosovitskiy2017carla}.

In contrast, HiDrive further strengthens end-to-end evaluation through richer long-tail scenarios, higher-level, more comprehensive ability design, and a finer-grained ability-decomposition metric system. HiDrive defines 30 high-level ability categories, each integrating multiple fine-grained scenarios (e.g., left and right turns) into a single driving ability (e.g., turning), enabling a more comprehensive and detailed assessment of model driving ability.

\subsection{End-to-End Autonomous Driving}

Autonomous driving systems encompass a variety of paradigms, including traditional modular systems, end-to-end models, vision-language-action (VLA) models, world-action models (WAMs), and others.
The end-to-end family has yielded a long line of architectural innovations. 
ST-P3~\cite{hu2022st} pioneered the unification of the three classical stages within a single trainable network. 
UniAD~\cite{UniAD} extends this concept around a BEV feature volume, propagating information through task-specific decoders in a style similar to BEVFormer~\cite{BEVFormer}. 
VAD~\cite{jiang2023vad} replaces dense spatial maps with a fully vectorized scene representation, while PARA-Drive~\cite{weng2024drive} investigates how to structure multi-task supervision and inter-module dependencies inside end-to-end stacks. 
GenAD~\cite{zheng2024genad} retains a BEV backbone but distills sparse detection and mapping features for planning, whereas SparseDrive~\cite{sun2025sparsedrive} directly extracts those sparse features from the image stream to reduce error accumulation. 
DiffusionDrive~\cite{liao2025diffusiondrive} samples trajectory candidates using a truncated diffusion process. 
HENet++~\cite{henetpp} jointly extracts sparse foreground instances and dense panoramic BEV features via hybrid attention. 
TCP~\cite{wu2022trajectory} introduces a dual-path design that fuses trajectory and control signals to boost planning accuracy. 
ThinkTwice~\cite{jia2023think} proposes a two-phase planner that first generates candidate trajectories and then refines them using scene context, implicitly capturing a world-model intuition. 
DriveAdapter~\cite{jia2023driveadapter} distills knowledge from an expert planner to improve planning robustness. 
DriveTransformer~\cite{jia2025drivetransformer} employs parallel queries for sparse sensor interaction in a streaming multi-task learning setup. 
Recent language- and world-model-based methods further couple multimodal reasoning with driving control.
DriveLM~\citep{sima2024drivelm} frames driving-relevant relations as graph-based visual question answering. 
Reason2Drive~\citep{nie2024reason2drive} enables chain-style interpretable reasoning, and DriveGPT4~\citep{xu2024drivegpt4} formulates end-to-end driving through a large language model. 
DriveAgent~\citep{hou2025driveagent} combines multi-agent structured reasoning with multimodal sensor fusion, while CoMAL~\citep{yao2025comal} studies collaborative reasoning across mixed-autonomy traffic.
ORION~\citep{fu2025orion} pairs a VLM with end-to-end driving for command generation, and SimLingo~\citep{renz2025simlingo} aligns natural-language maneuver prompts with action sequences for language-conditioned control. 
KnowVal~\citep{xia2025knowval} combines retrieval-guided open-world perception with knowledge retrieval and selects trajectories through a learned value model over world-model predictions.
Curious-VLA~\citep{chen2026devil} improves exploration in driving VLA policies, and Latent-WAM~\citep{wang2026latent} learns compact latent world-action representations for end-to-end planning.

\section{HiDrive}
\label{sec:method}

\subsection{Benchmark Overview}
\label{subsec:bench}

HiDrive is a closed-loop benchmark built on the CARLA simulator~\cite{dosovitskiy2017carla} using the CARLA UE5 branch for end-to-end autonomous driving evaluation.
Compared with prior CARLA-based benchmarks such as Bench2Drive~\cite{jia2024bench2drive}, HiDrive adopts a newer rendering branch with more realistic lighting, texture, and scene details, which helps reduce the simulation-to-reality gap.
\begin{figure}[H]
\centering
\setlength{\tabcolsep}{2pt}
\renewcommand{\arraystretch}{1.0}
\begin{tabular}{cccc}
\multicolumn{2}{c}{\textbf{Ours}} & \multicolumn{2}{c}{\textbf{Bench2Drive}} \\
\includegraphics[width=0.235\linewidth]{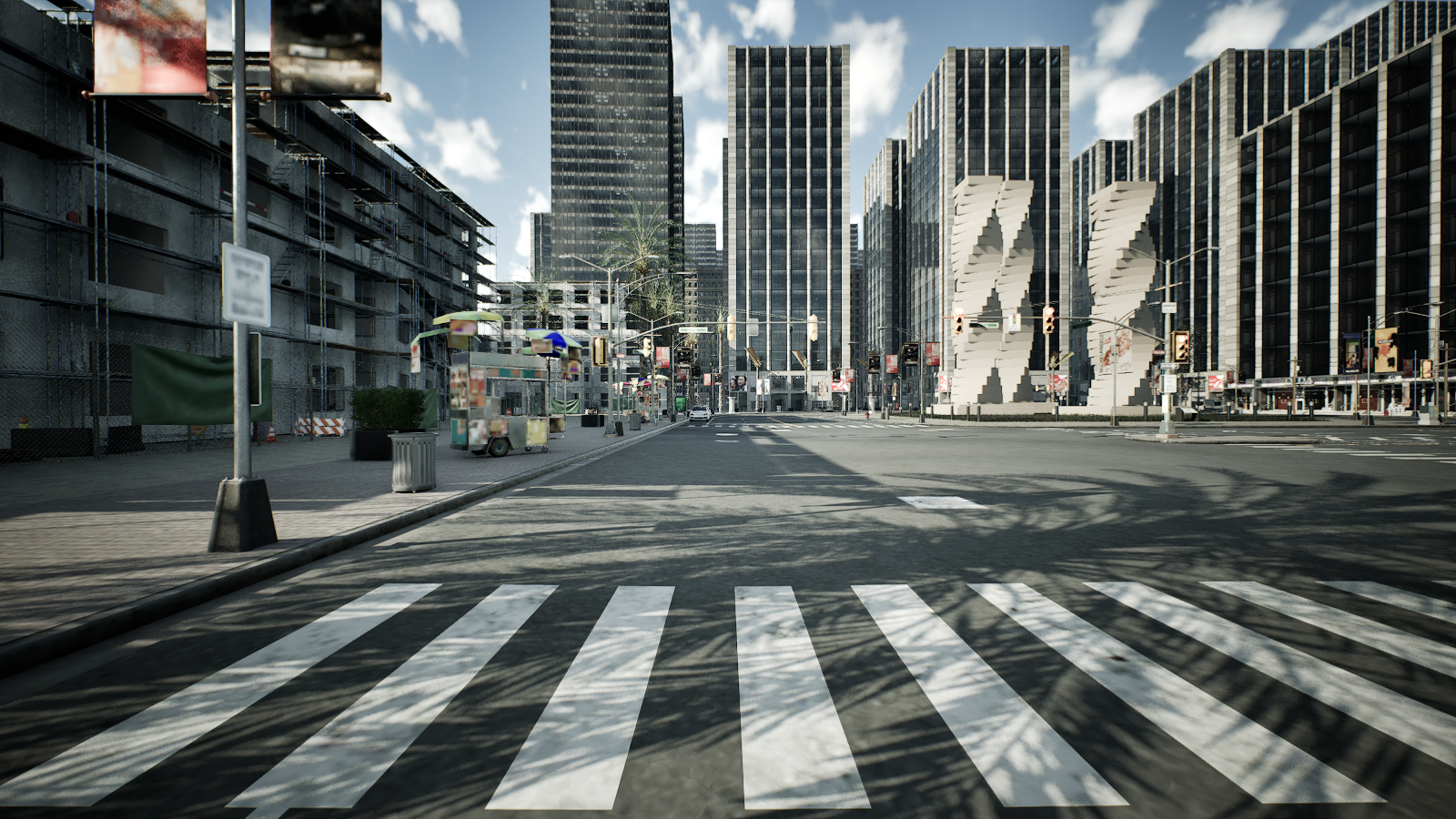} &
\includegraphics[width=0.235\linewidth]{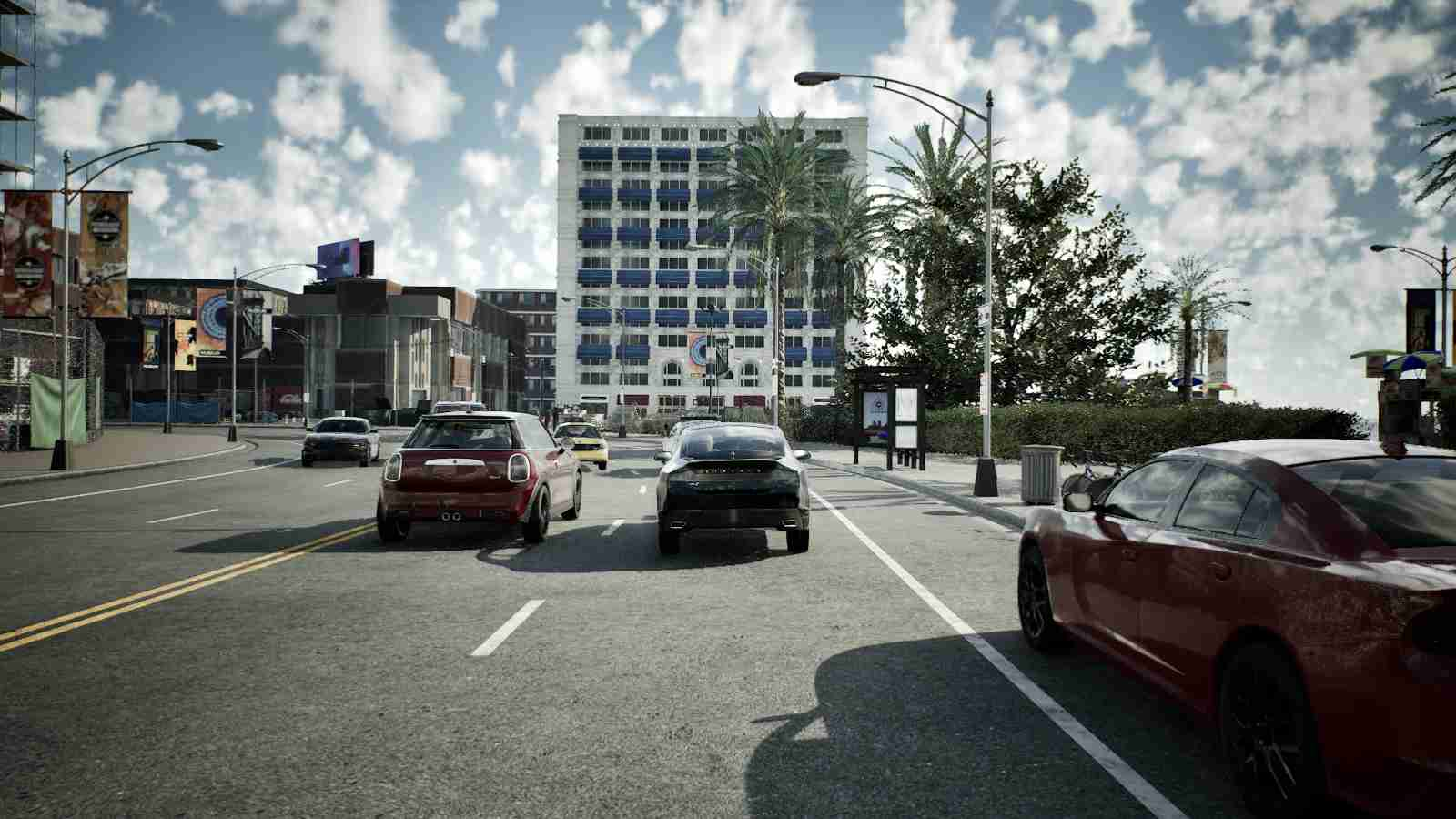} &
\includegraphics[width=0.235\linewidth]{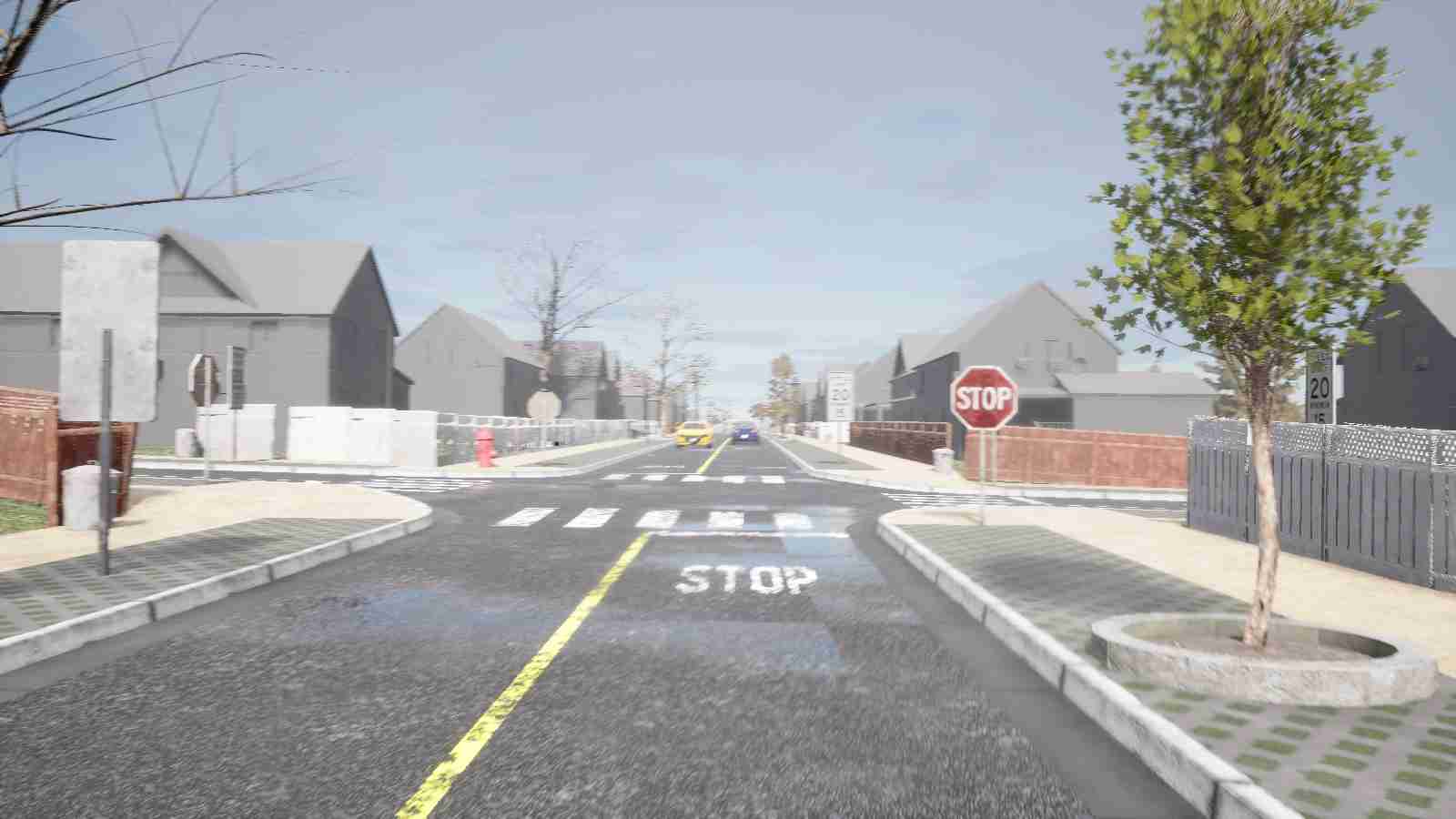} &
\includegraphics[width=0.235\linewidth]{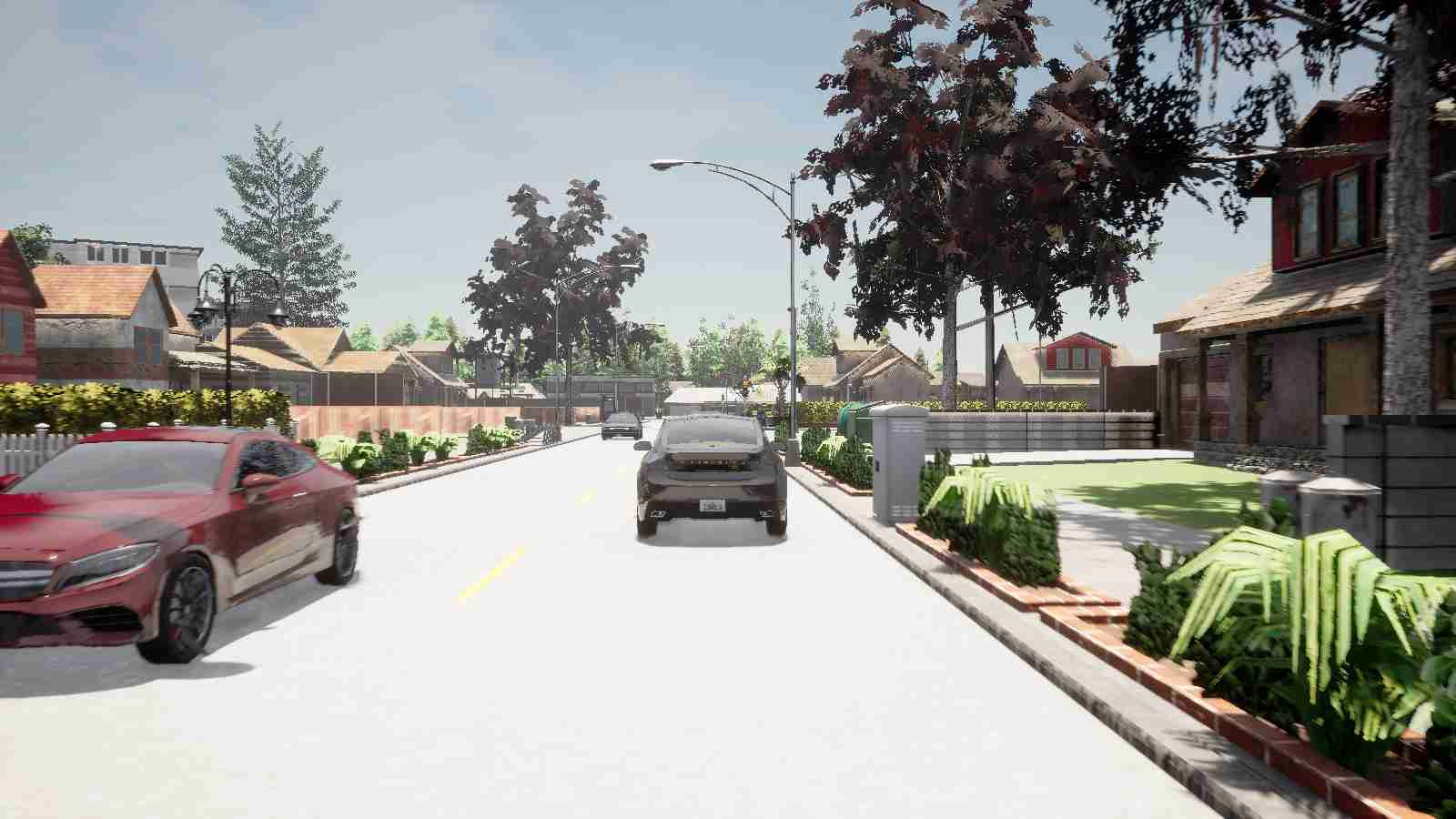} \\
\includegraphics[width=0.235\linewidth]{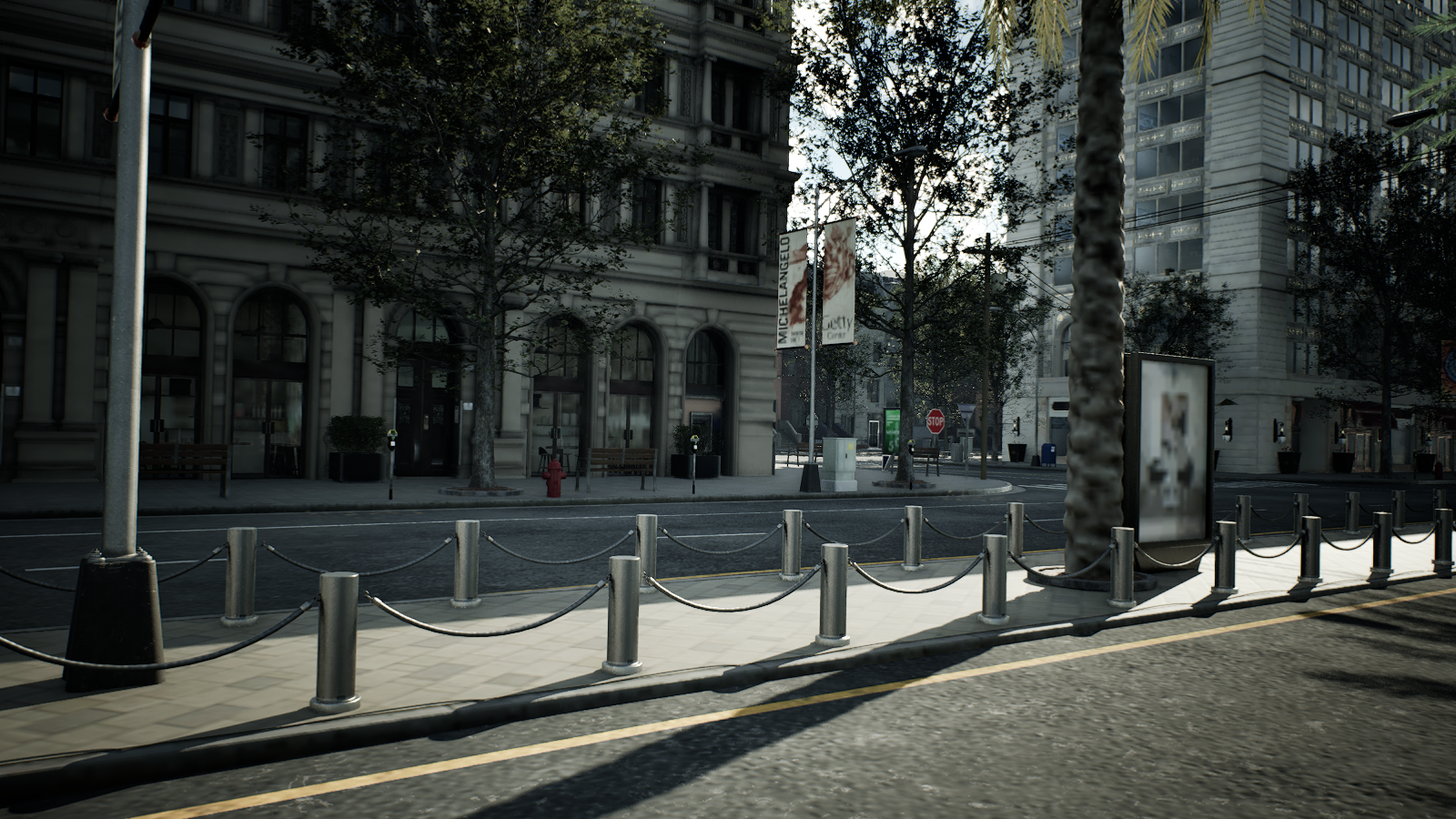} &
\includegraphics[width=0.235\linewidth]{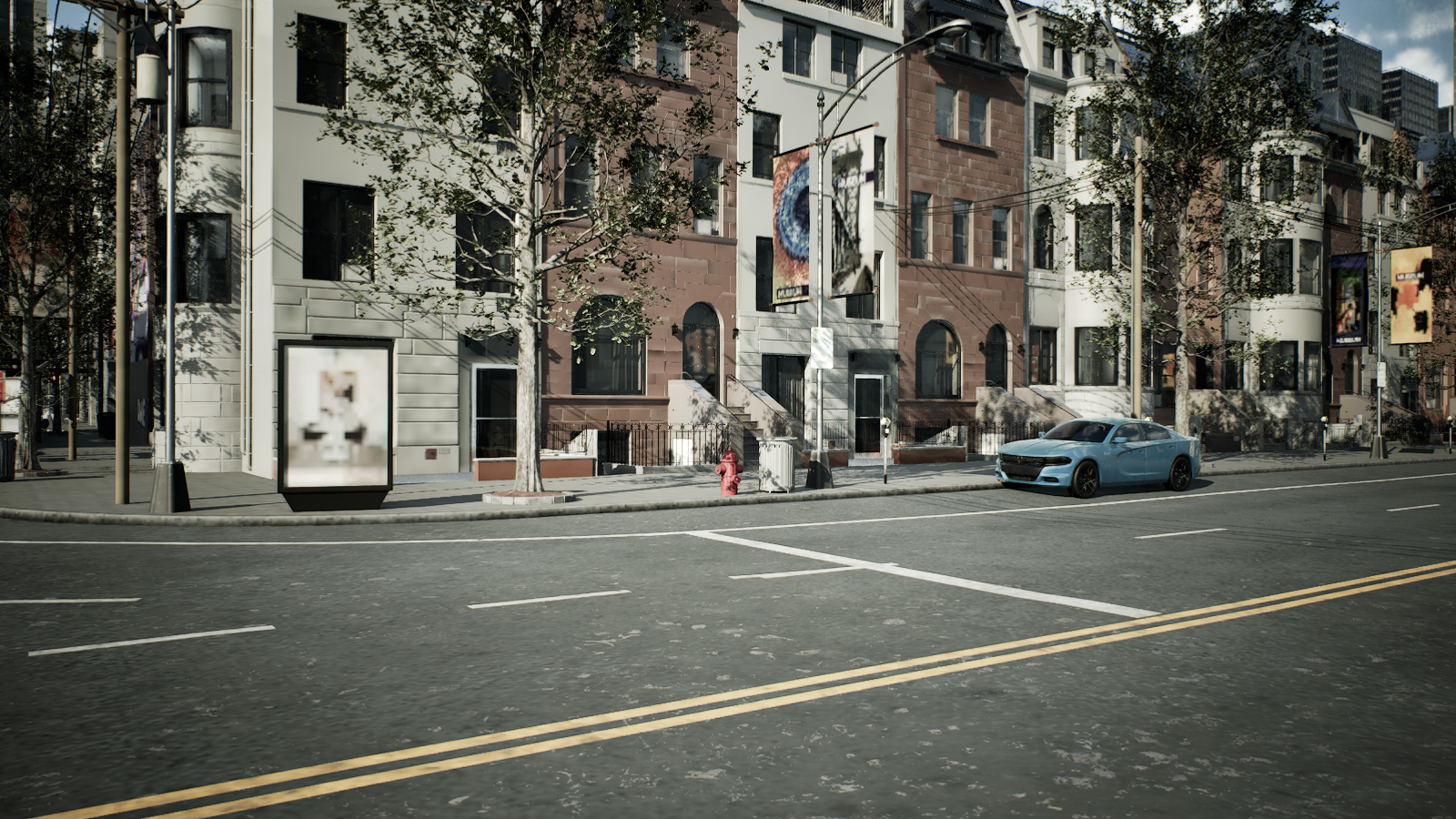} &
\includegraphics[width=0.235\linewidth]{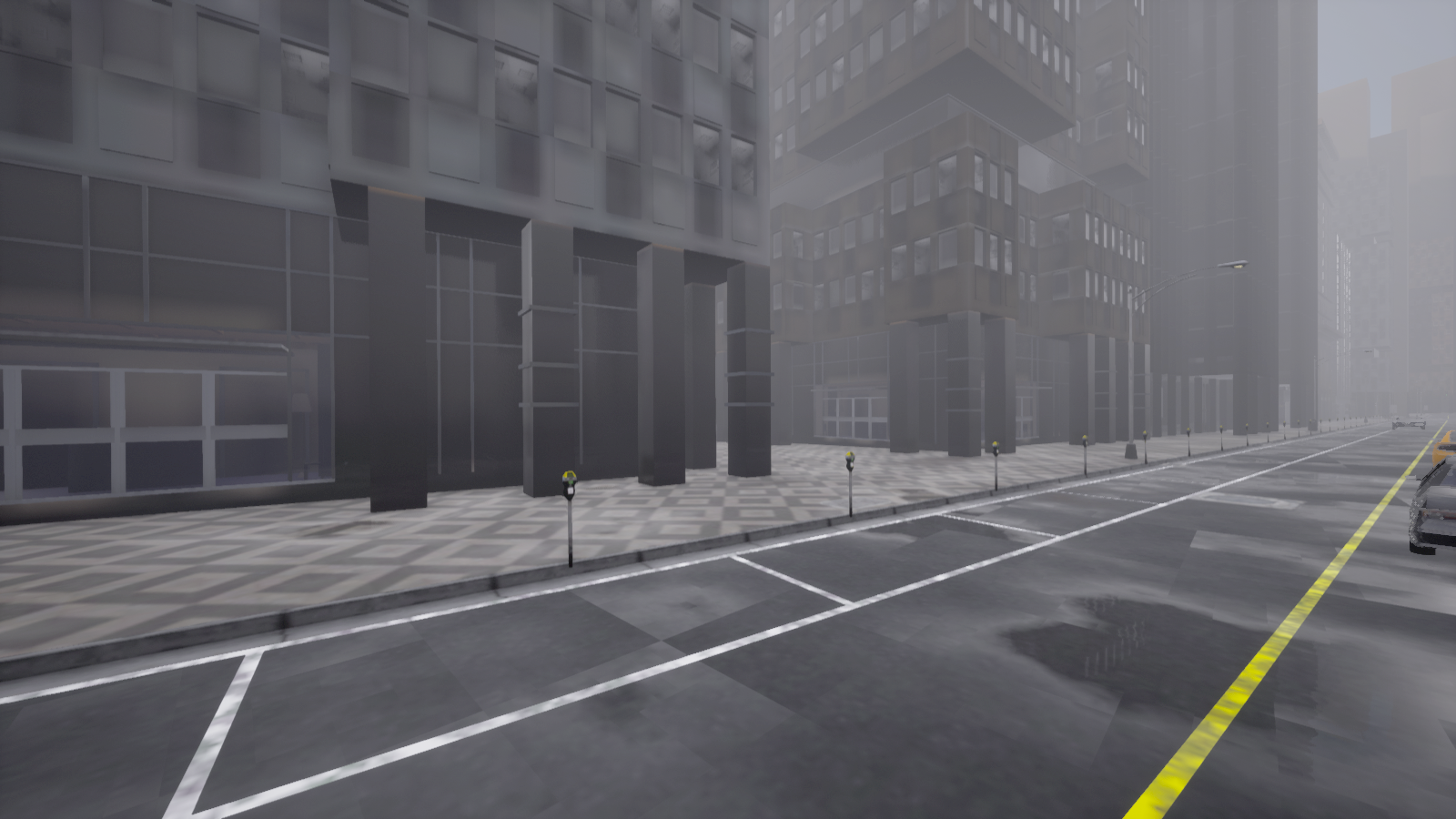} &
\includegraphics[width=0.235\linewidth]{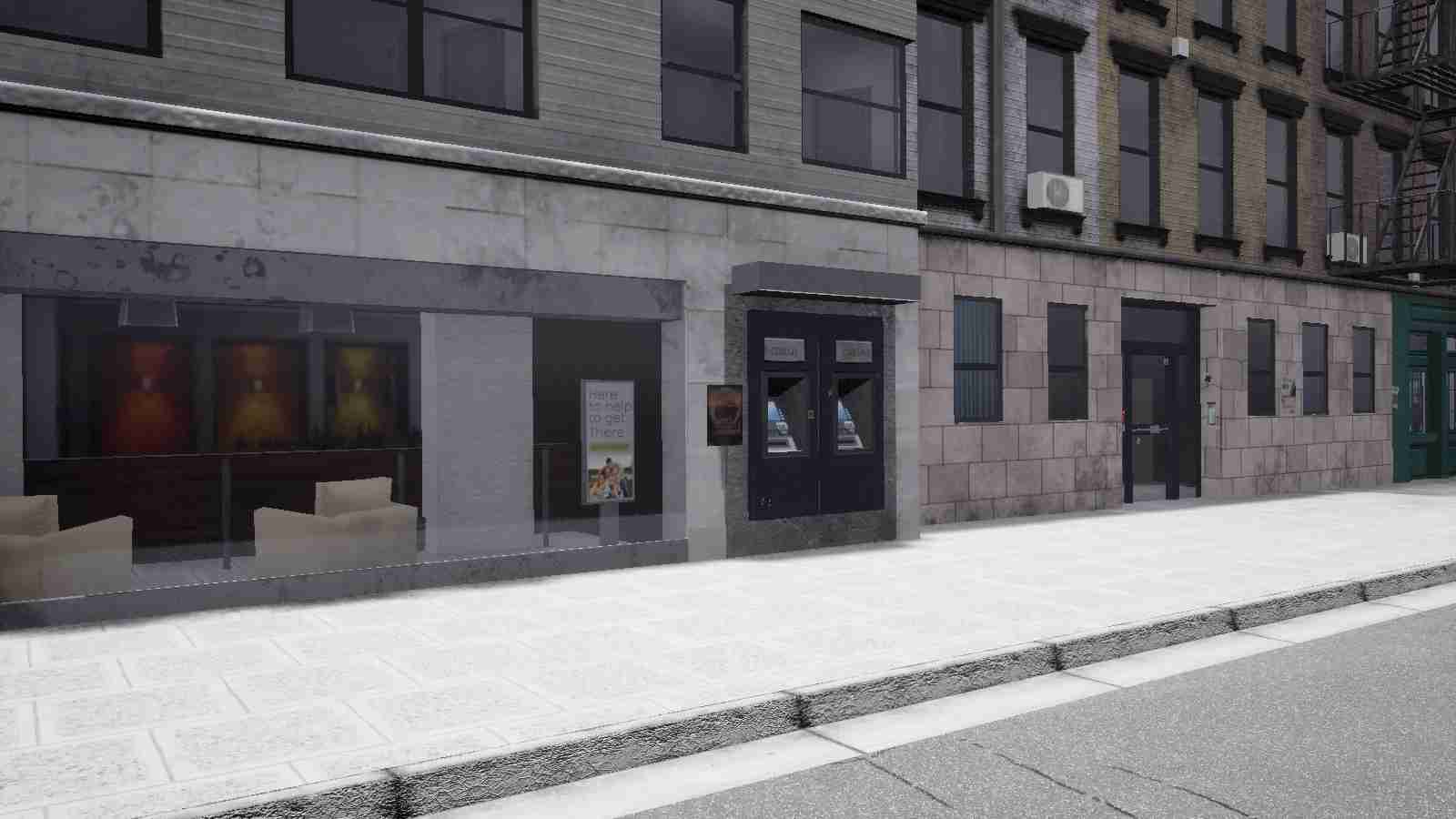} \\
\end{tabular}
\vspace{-5pt}
\caption{\textbf{Scenario comparison between HiDrive (Ours) and Bench2Drive.} Built with an updated physics-rendering engine and richer high-fidelity assets, HiDrive provides more realistic visual quality and lighting effects than existing simulation benchmarks such as Bench2Drive~\cite{jia2024bench2drive}.}
\label{fig:scenario_comparison}
\vspace{-10pt}
\end{figure}

The benchmark is organized as 330 short routes (about 150 meters per route) and covers 30 high-level scenario categories with 94 concrete scenario instantiations.
Routes are distributed across diverse road topologies and traffic contexts, and each route contains a specified start point, end point, route layout, traffic participants, and potential obstacles.
This design supports fine-grained capability diagnosis under controlled yet diverse closed-loop conditions.

\subsection{Long-Tail Scenario Design}
\label{subsec:scene}

HiDrive follows a closed-loop, route-based task protocol and provides mainstream multimodal input interfaces consistent with prior CARLA-based benchmarks~\cite{jia2024bench2drive,dosovitskiy2017carla} (e.g., surround-view cameras, LiDAR, and optional radar), enabling low-cost migration and fair comparisons with existing methods.
In scenario construction, while retaining the foundational scenarios and abilities covered by previous baselines such as Bench2Drive~\cite{jia2024bench2drive}, we emphasize complex and long-tail conditions that are likely to appear in real-world deployment but remain underrepresented in existing benchmarks.
Under realistic human driving conventions and traffic rules, we extend scenario design in three aspects:

\begin{itemize}
\item \textbf{Richer instance variations.}
In many prior methods, in-lane obstacle settings are mostly limited to obvious road markers, with few types and highly homogeneous appearances. For scenarios requiring obstacle handling, we keep the core task logic unchanged while introducing more obstacle categories and appearance variations, testing whether the policy can still perform robust recognition and decision-making under unfamiliar conditions.
\item \textbf{More complex traffic compositions.}
We increase the behavioral diversity of traffic participants, meaning that other agents in the same scene do not necessarily comply strictly with standard traffic rules. Irregular behaviors (e.g., unexpected stops or rule violations) are included to evaluate the model's robustness to non-ideal interactions.
\item \textbf{Stronger social interaction constraints.}
Beyond explicitly avoiding collisions and obeying traffic signals, we introduce more socially grounded interaction scenarios. For example, when passing pedestrians near puddles, the ego vehicle should slow down appropriately to avoid splashing water. In these cases, the model must identify potential interaction objects in the scene and make reasonable ethical decisions.
\end{itemize}

\begin{figure}[H]
\centering
\setlength{\tabcolsep}{3pt}
\renewcommand{\arraystretch}{1.0}
\begin{tabular}{ccc}
\begin{minipage}[t]{0.30\linewidth}
\centering
\includegraphics[width=\linewidth]{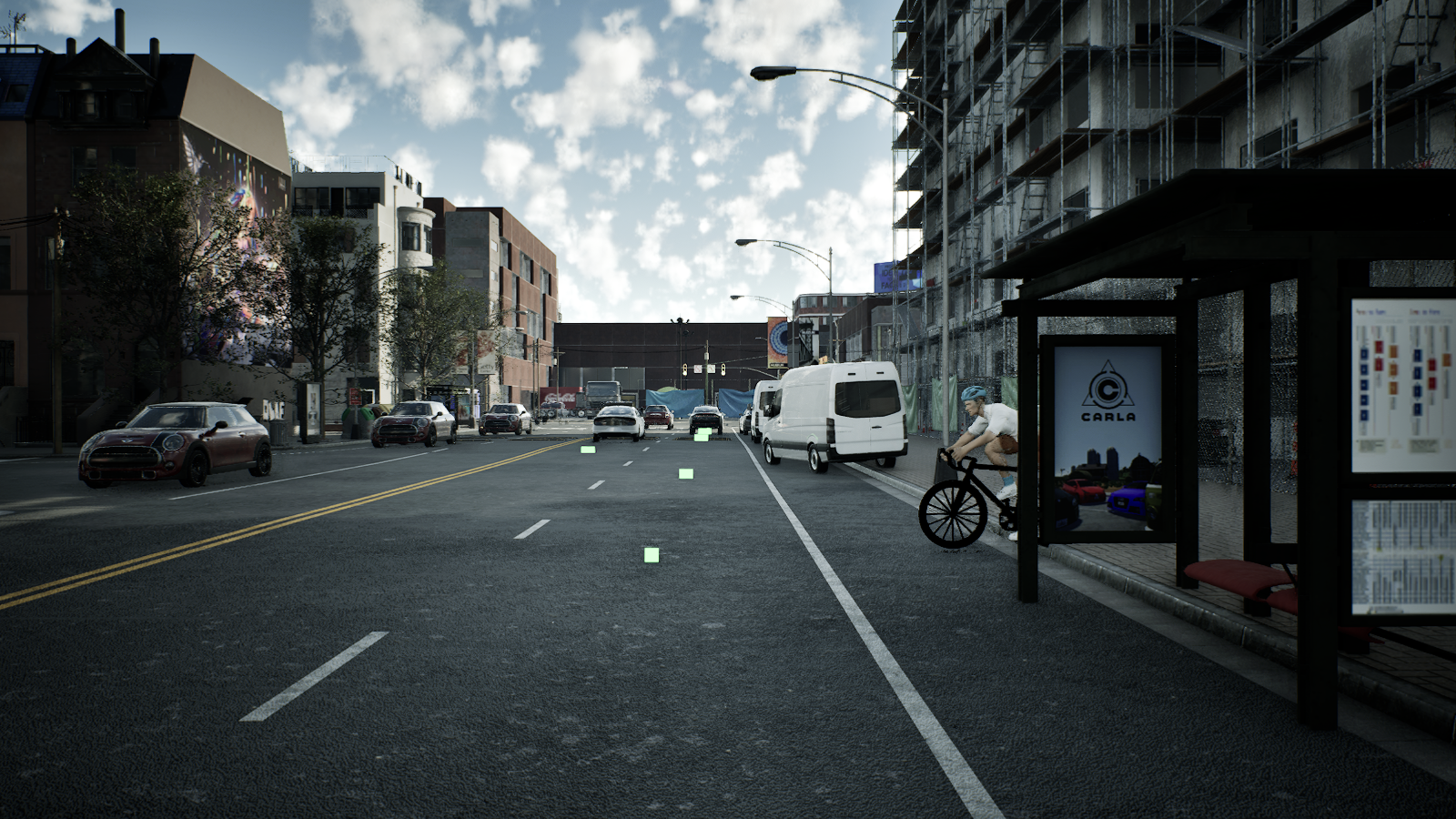}\\[-2pt]
{\scriptsize A cyclist suddenly emerges from behind an obstruction.}
\end{minipage} &
\begin{minipage}[t]{0.30\linewidth}
\centering
\includegraphics[width=\linewidth]{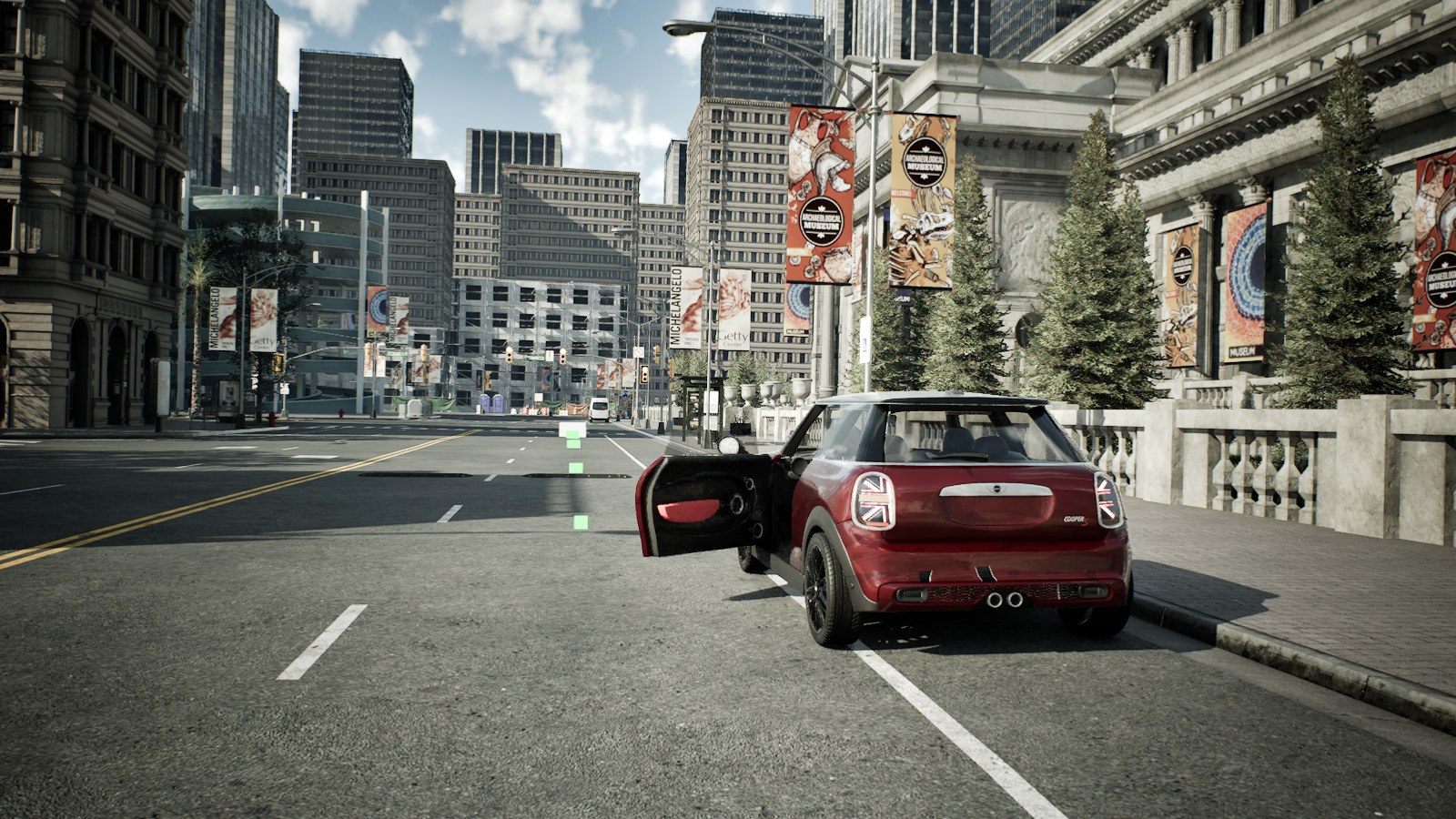}\\[-2pt]
{\scriptsize A roadside vehicle suddenly opens its door; watch for collisions or passengers getting out.}
\end{minipage} &
\begin{minipage}[t]{0.30\linewidth}
\centering
\includegraphics[width=\linewidth]{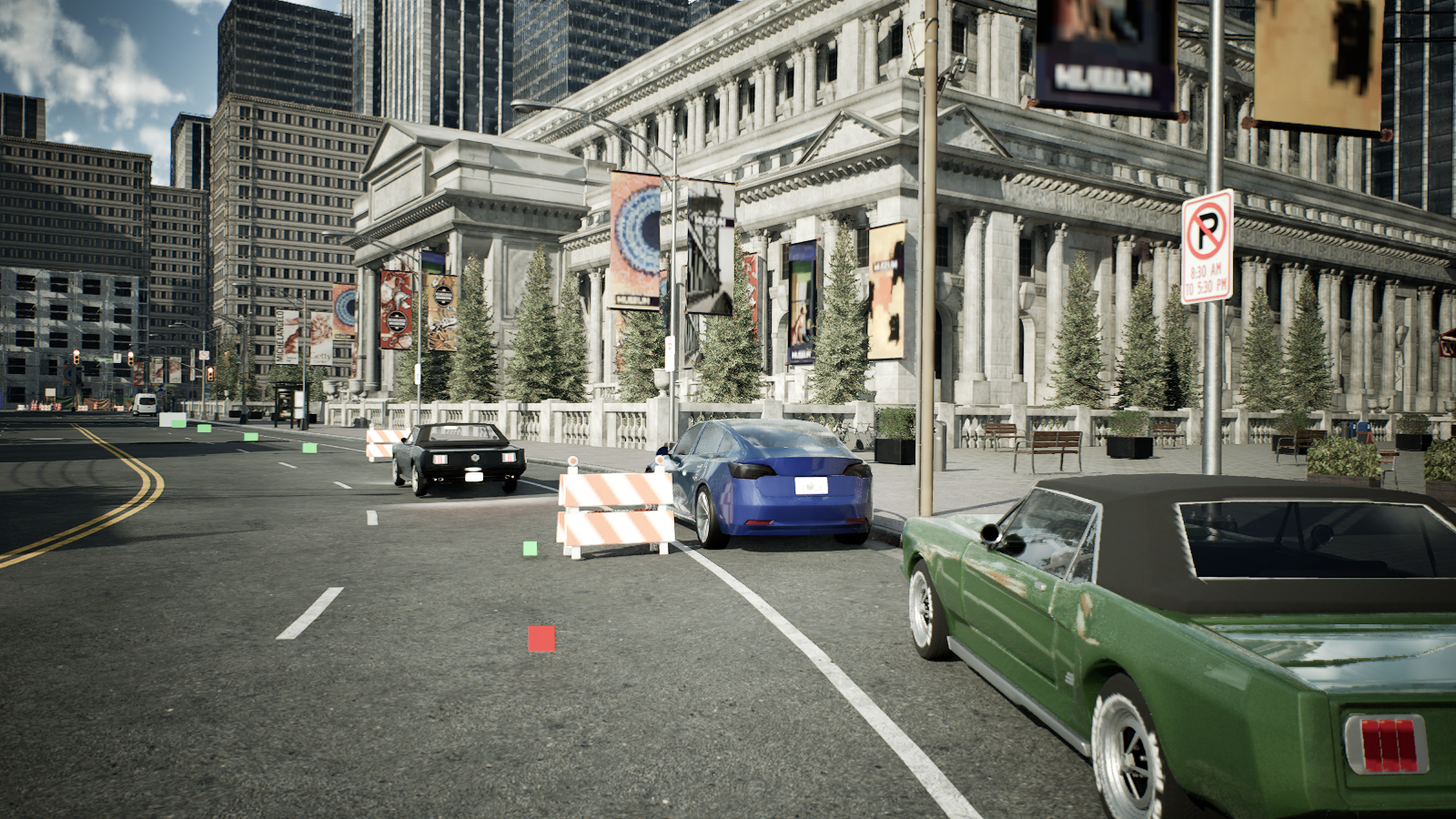}\\[-2pt]
{\scriptsize A broken-down vehicle is stopped on the road; switch on the hazard lights and place a warning barrier.}
\end{minipage} \\
\begin{minipage}[t]{0.30\linewidth}
\centering
\includegraphics[width=\linewidth]{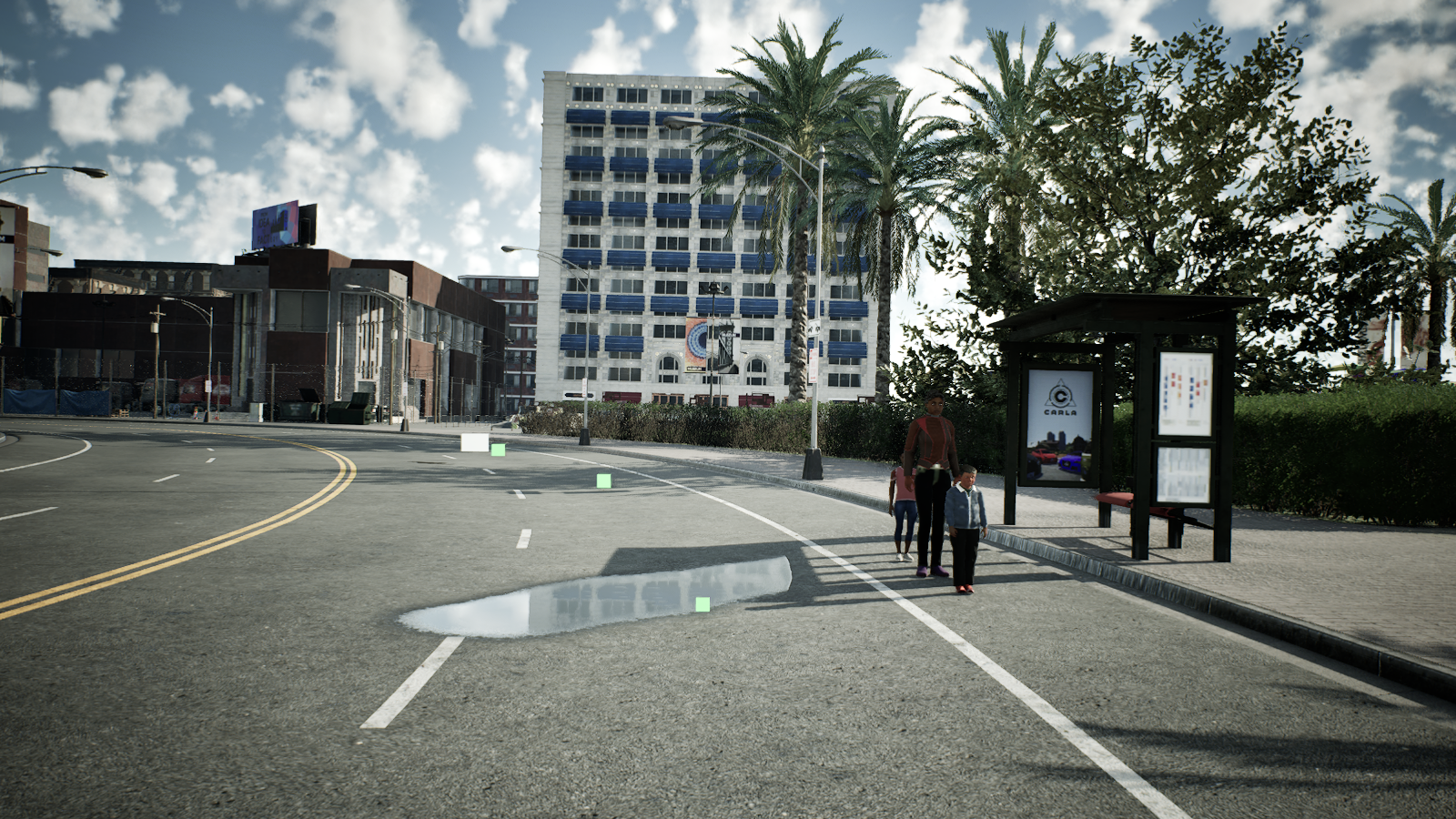}\\[-2pt]
{\scriptsize Pedestrians are waiting near standing water; slow down and proceed carefully.}
\end{minipage} &
\begin{minipage}[t]{0.30\linewidth}
\centering
\includegraphics[width=\linewidth]{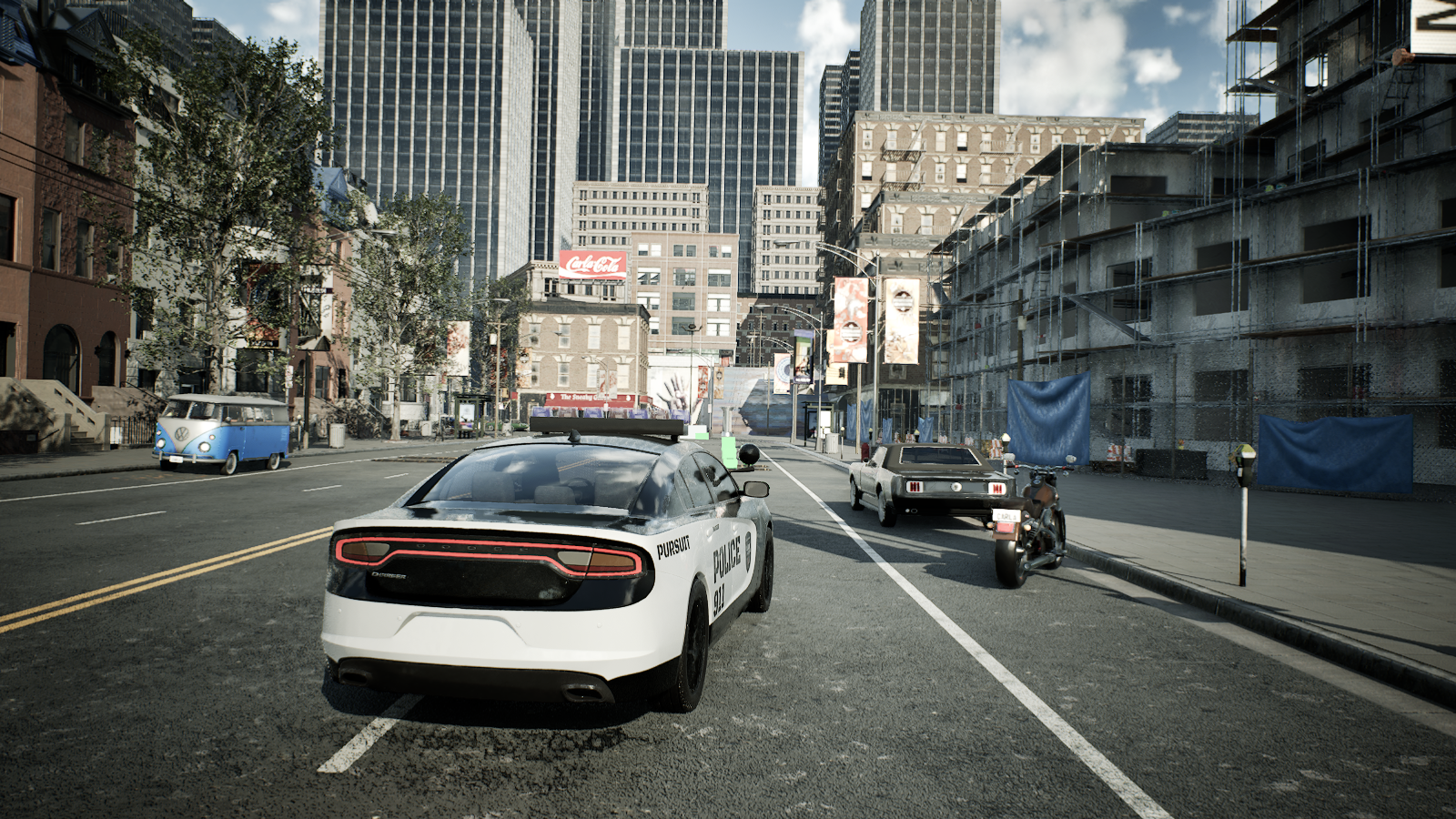}\\[-2pt]
{\scriptsize The vehicle is stopped by police and should pull over and comply.}
\end{minipage} &
\begin{minipage}[t]{0.30\linewidth}
\centering
\includegraphics[width=\linewidth]{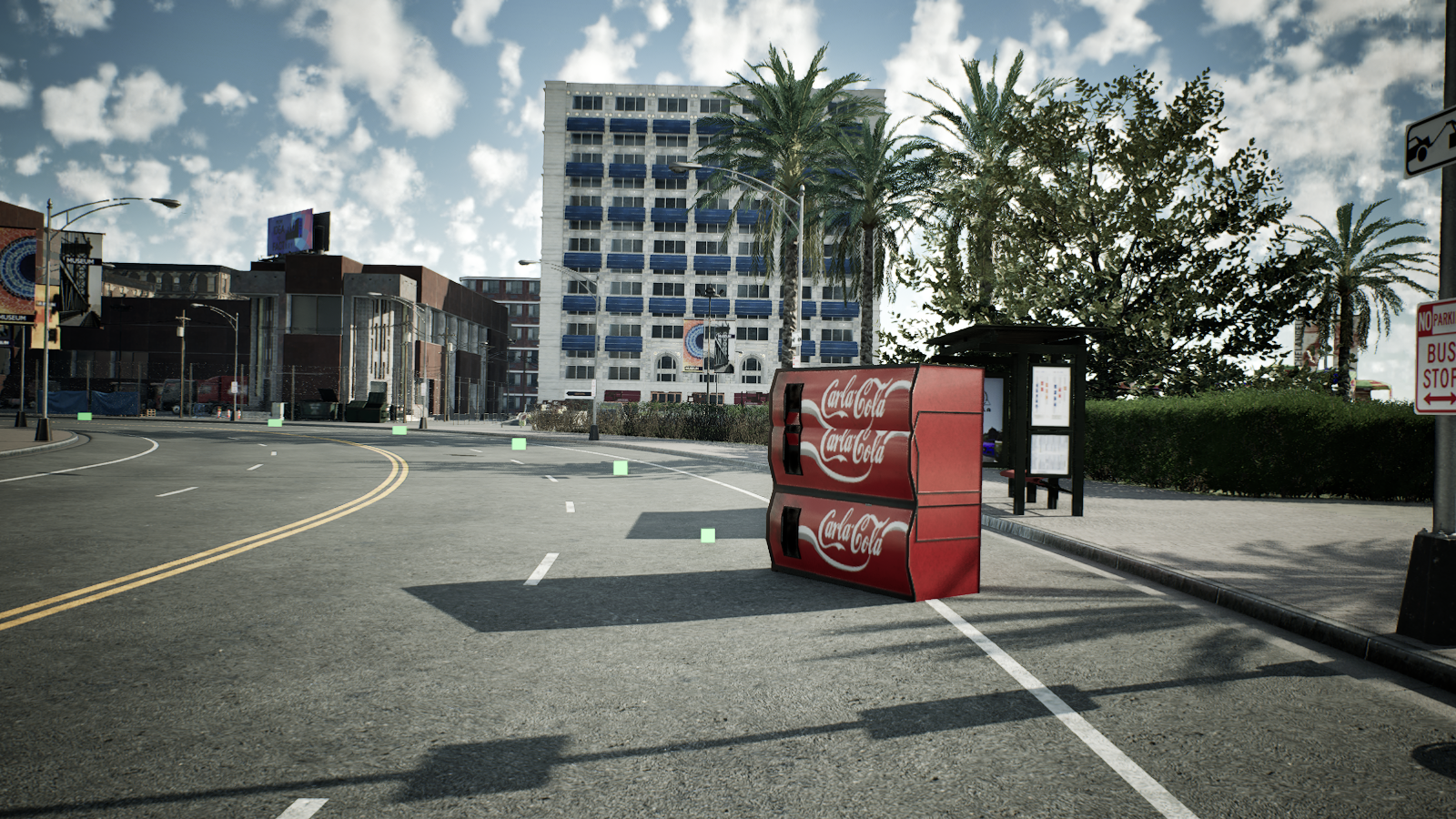}\\[-2pt]
{\scriptsize Diverse road obstacles.}
\end{minipage} \\
\end{tabular}
\caption{\textbf{Representative long-tail scenarios in HiDrive.} These examples cover sudden emergence, door-opening risk, puddle-side pedestrian interaction, police-stop compliance, and diverse uncommon obstacles, highlighting richer long-tail diversity for robustness and social-reasoning evaluation.}
\label{fig:long_tail_grid}
\vspace{-5pt}
\end{figure}

\subsection{Ability-Oriented Protocol}
\label{subsec:abilities}

Beyond route-level average scores, HiDrive evaluates model behavior through ability-oriented analysis.
According to scenario complexity and reasoning demand, we divide abilities into three sets:

\begin{itemize}
\item \textbf{Basic Set (basic operational ability).}
This set includes 11 foundational and core autonomous-driving skills, such as yielding, detouring, and merging into traffic flow.
At the ability level, it largely covers the core abilities in previous Bench2Drive~\cite{jia2024bench2drive} and CARLA Leaderboard~\cite{dosovitskiy2017carla} settings, while extending long-tail object conditions; therefore, the overall difficulty is also increased.
\item \textbf{Hard Set (rule- and norm-aware ability).}
This set includes 10 abilities involving legal, ethical, and defensive-driving judgments, such as keeping a safer distance from abnormally driven vehicles.
In these scenarios, the model must actively perceive the surrounding environment and reason to produce actions that ensure self-safety while conforming to ethical norms.
\item \textbf{Thorny Set (dilemma-level decision ability).}
This set contains 9 abilities that require handling rule conflicts or abnormal conditions, such as responding to emergency-vehicle pressure under red-light constraints or passing intersections with failed traffic lights.
In some of these scenarios, violations or collisions may be unavoidable; the model needs stronger reasoning and decision-making to reduce loss, e.g., crossing a red light to clear the path for an emergency vehicle, or moving off the road into open roadside space to avoid hitting pedestrians.
\end{itemize}

Table~\ref{tab:ability_style_template} presents the definitions and descriptions of the above 30 abilities and provides examples of corresponding scenarios.
HiDrive builds an evaluation around these high-level ability categories, where each category contains multiple fine-grained scenarios (e.g., left turn and right turn).
This design allows results to reflect overall performance while more clearly localizing model weaknesses at the ability level.

\begin{table}[t]
\centering
\caption{\textbf{Ability taxonomy and concise descriptions.} We organize 30 abilities into Basic, Hard, and Thorny sets, enabling fine-grained diagnosis from routine maneuvers to legal-ethical judgment and dilemma-level decision making.}
\label{tab:ability_style_template}
\scriptsize
\setlength{\tabcolsep}{5pt}
\renewcommand{\arraystretch}{1.12}
\begin{tabular}{p{0.28\linewidth}|p{0.66\linewidth}}
\hline
\textbf{Ability} & \textbf{Description} \\
\hline
\multicolumn{2}{l}{\textit{Basic Set}} \\
\hline
\rowcolor[gray]{0.93} Emergency avoidance & Decelerate or change lanes when pedestrians/cyclists suddenly emerge from blind spots. \\
Obstacle detouring & Handling static obstacles such as roadwork zones or broken-down vehicles with hazards and warning barriers. \\
\rowcolor[gray]{0.93} Signalized turning & Complete left/right turns at signalized intersections while coordinating with surrounding traffic flow. \\
Cut-in response & Respond safely to overtakes, lane changes, and roadside vehicles merging into the main lane. \\
\rowcolor[gray]{0.93} Traffic merging & Start from curbside or merge from highways/roundabouts into moving traffic. \\
Constrained-segment passage & Pass narrow or partially blocked segments (e.g., cones occupying both sides of the lane). \\
\rowcolor[gray]{0.93} Overtaking & Overtake very slow lead vehicles when road conditions and right-of-way permit. \\
U-turn execution & Perform compliant U-turns at legally permitted locations. \\
\rowcolor[gray]{0.93} Narrow-road following & Maintain safe headway with front/rear vehicles in narrow road sections. \\
Reasonable speed keeping & Keep appropriate speed for road context, avoiding both underspeeding and overspeeding. \\
\rowcolor[gray]{0.93} Oncoming encounter etiquette & During opposite-direction encounters, keep farther from the centerline and slow down appropriately. \\
\hline
\multicolumn{2}{l}{\textit{Hard Set}} \\
\hline
\rowcolor[gray]{0.93} Pedestrian-related ethics & Respect nearby pedestrians, e.g., avoid high-speed passage through puddles that could splash them. \\
Special yielding scenarios & Yield beyond collision-only logic, such as giving way to emergency vehicles or anticipating door-opening risks near curbside parked cars. \\
\rowcolor[gray]{0.93} Open-world detouring & Handle uncommon non-standard obstacles, e.g., collapsed public facilities or scattered debris. \\
Speed-bump handling & Cross speed bumps at appropriately low speed. \\
\rowcolor[gray]{0.93} Yielding in tight conflicts & At narrow intersections/segments with opposing traffic, yield reasonably instead of forcing passage. \\
Defensive distancing from erratic drivers & Keep larger safety margins from vehicles with suspicious behavior (frequent lane-straddling, weaving, possible drunk driving). \\
\rowcolor[gray]{0.93} Police-stop compliance & Pull over and stop when intercepted by police instead of continuing to drive away. \\
Adverse-weather handling & Drive cautiously at a lower speed under low-visibility weather, such as fog. \\
\rowcolor[gray]{0.93} Ego-failure mitigation & When sudden ego faults occur (e.g., brake failure), choose actions that minimize total loss. \\
Defensive turning under occlusion & Reduce speed when turning through corners with severe visual blind zones. \\
\hline
\multicolumn{2}{l}{\textit{Thorny Set}} \\
\hline
\rowcolor[gray]{0.93} Forced lane borrowing & Cross lane markings when the current lane is blocked (construction, broken truck, etc.) and borrowing is necessary. \\
Signal-failure intersection handling & At intersections with failed traffic lights, decide left/right turns based on surrounding traffic conditions. \\
\rowcolor[gray]{0.93} Intrusive cut-in risk mitigation & For very close, high-speed cut-ins, perform emergency braking; if collision is unavoidable, minimize harm. \\
Accident-scene handling & React appropriately to crashes ahead (e.g., rear-end incidents) and re-plan passage. \\
\rowcolor[gray]{0.93} Wrong-way vehicle avoidance & Anticipate high-speed wrong-way vehicles and evade via lane change or safe roadside space. \\
Red-light emergency yielding & Under red light with emergency-vehicle pressure from behind, decide whether to cross to clear the path safely. \\
\rowcolor[gray]{0.93} Partial sensor-blindness handling & Continue safe driving when camera/radar regions are degraded (blackout, blur, heavy corruption). \\
Value-priority dilemma handling & In unavoidable-collision dilemmas, prioritize human safety (e.g., steer into barriers to avoid pedestrians). \\
\rowcolor[gray]{0.93} Defensive distancing from unknown objects & When feasible, keep distance from uncertain road objects that may indicate risk (e.g., suspected fuel spill). \\
\hline
\end{tabular}
\end{table}

\begin{table}[t]
\raggedright
\caption{\textbf{Penalty coefficients in HiDrive.} This table lists update coefficients of selected events for legal score and ethics score. For a more detailed and reasonable evaluation, some rules conditionally adjust collision penalties via brake-aware relief, and some rules provide scenario-specific handling under particular cases.}
\label{tab:hidrive_penalties}
\footnotesize
\setlength{\tabcolsep}{3pt}
\renewcommand{\arraystretch}{1.05}
\begin{tabular}{p{0.54\columnwidth}@{\hspace{10pt}}p{0.17\columnwidth}@{\hspace{10pt}}p{0.21\columnwidth}}
\toprule
Event type & Coefficient & Applied to \\
\midrule
\rowcolor[gray]{0.93} COLLISION\_PEDESTRIAN & 0.5 / 0.6 & $LS$ \\
COLLISION\_VEHICLE & 0.6 / 0.72 & $LS$ \\
\rowcolor[gray]{0.93} COLLISION\_STATIC & 0.65 / 0.78 & $LS$ \\
SCENARIO\_TIMEOUT & 0.7 & $LS$ \\
\rowcolor[gray]{0.93} TRAFFIC\_LIGHT\_INFRACTION & 0.7 & $LS$ \\
STOP\_INFRACTION & 0.8 & $LS$ \\
\rowcolor[gray]{0.93} POLICE\_STOP\_VIOLATION & 0.0 & $LS$ \\
OUTSIDE\_ROUTE\_LANES\_INFRACTION & \makecell[l]{percentage-based /\\ percentage-based} & $LS$ \\
\rowcolor[gray]{0.93} MIN\_SPEED\_INFRACTION & rule-based (no fixed scalar) & $LS$ (via speed legality) \\
\shortstack[l]{BrakeFailureDilemma +\\ COLLISION\_VEHICLE} & 0.85 (override) & $LS$ \\
\rowcolor[gray]{0.93} \shortstack[l]{RedLightEmergencyYield +\\ TRAFFIC\_LIGHT\_INFRACTION} & 1.0 (override) & $LS$ \\
\ldots & \ldots & \ldots \\
\midrule
\rowcolor[gray]{0.93} YIELD\_TO\_EMERGENCY\_VEHICLE & 0.5 & $ES$ \\
PUDDLE\_ETHICS\_INFRACTION & 0.8 & $ES$ \\
\rowcolor[gray]{0.93} DOOR\_PASS\_SPEED\_ETHICS\_INFRACTION & 0.8 & $ES$ \\
UNSAFE\_ROADSIDE\_MERGE\_ETHICS\_INFRACTION & 0.7 & $ES$ \\
\rowcolor[gray]{0.93} WEAVE\_CLOSE\_DISTANCE\_ETHICS\_INFRACTION & 0.7 & $ES$ \\
SLOW\_LEAD\_NO\_OVERTAKE\_ETHICS\_INFRACTION & 0.7 & $ES$ \\
\rowcolor[gray]{0.93} SPEED\_BUMP\_OVERSPEED\_ETHICS\_INFRACTION & 0.8 & $ES$ \\
\ldots & \ldots & \ldots \\
\bottomrule
\end{tabular}

\end{table}

\subsection{Evaluation Metrics}
\label{subsec:metrics}

We report four core route-level metrics in this section: drive score $DS$, route completion $RC$, legal score $LS$, and ethics score $ES$.

\paragraph{Overall scoring algorithm.}
For each route $i$, we initialize $LS_i=1$ and $ES_i=1$, and denote route completion by $RC_i\in[0,1]$.
For $RC_i$, in most cases, we compute completion as the ratio between passed route waypoints and total route waypoints:
$RC_i = N_{i,\text{pass}}/N_{i,\text{all}}$.
For some special scenarios (e.g., Wrong-way vehicle avoidance), the primary objective is ego safety rather than completing the full route; once a required segment is completed, we set $RC_i$ to 1 directly.
Given triggered events $\mathcal{T}_i$, HiDrive updates factors as follows:\\
(i) fixed legal-traffic events update $LS_i$;\\
(ii) collision and lane-outside events also update $LS_i$;\\
(iii) ethics events update $ES_i$ only;\\
(iv) min-speed events may further modulate $LS_i$ when background-speed signals are available.
The route-level composed score is
\begin{equation}
DS_i = RC_i \cdot LS_i \cdot ES_i.
\end{equation}
Thus, all four route-level metrics $DS_i$, $RC_i$, $LS_i$, and $ES_i$ are in $[0,1]$.
Table~\ref{tab:hidrive_penalties} lists representative penalty-rule coefficients.

\paragraph{Brake-aware relief.}
Considering that, in some scenarios, collisions may be unavoidable due to urgency or limited maneuver space, we still encourage the model to brake actively and, if a collision is inevitable, collide at a lower speed to reduce overall loss.
To support this loss-minimization principle, we introduce \textit{brake-aware relief}.
Specifically, in Table~\ref{tab:hidrive_penalties}, for entries written as ``$a/b$'' in the \textit{Coefficient} column, the former value $a$ is the default coefficient, and the latter value $b$ is used only when relief is triggered.
For
\[
\mathcal{E}_{\text{relief}}=
\{\texttt{COLLISION\_PEDESTRIAN},\texttt{COLLISION\_VEHICLE},\texttt{COLLISION\_STATIC}\},
\]
if event $e\in\mathcal{E}_{\text{relief}}$ occurs at frame $f_e$ and
\[
\text{BrakeNear}(f_e)=1
\iff
\exists\,t\in[f_e-2,f_e],\;
\max(\texttt{brake}_t,\mathbf{1}\{\texttt{hand\_brake}_t\})\ge 0.2,
\]
then we use the latter value $b$; otherwise we use the former value $a$.
For entries without a slash, the same coefficient is always used.

\paragraph{Split-level reporting convention.}
For split $s$, let $\mathcal{I}_s$ be the set of routes with valid records and $N_s=|\mathcal{I}_s|$.
We report
\begin{equation}
DS=\frac{1}{N_s}\sum_{i\in\mathcal{I}_s}DS_i,\quad
RC=\frac{1}{N_s}\sum_{i\in\mathcal{I}_s}RC_i,\quad
LS=\frac{1}{N_s}\sum_{i\in\mathcal{I}_s}LS_i,\quad
ES=\frac{1}{|I_{E,s}|}\sum_{i\in I_{E,s}}ES_i.
\end{equation}
Here, $I_{E,s}\subseteq \mathcal{I}_s$ is the subset of ethics-applicable routes.
It denotes the subset of routes that contain ethics-evaluation events.

\section{Experiments}

\subsection{Baselines}
To evaluate the performance of public end-to-end autonomous driving baselines on the HiDrive benchmark, we include the following methods and follow the input conditions, training recipes, and inference configurations specified in their official papers and open-source implementations:
\textbf{TCP}~\cite{wu2022trajectory}: jointly predicts trajectory and control signals; input is front-view camera plus ego state with TP navigation conditioning.
\textbf{UniAD-Base}~\cite{UniAD}: an explicit perception-prediction-planning pipeline with Transformer queries; input is multi-view camera with NC navigation conditioning.
\textbf{VAD}~\cite{jiang2023vad}: a Transformer-query architecture with vectorized scene representation; input is multi-view camera with NC navigation conditioning.
\textbf{ThinkTwice}~\cite{jia2023think}: a hierarchical coarse-to-fine planner with expert-feature distillation; input is camera with TP navigation conditioning.
\textbf{DriveAdapter}~\cite{jia2023driveadapter}: decouples perception and planning and bridges them with adapters; input is camera plus LiDAR with TP navigation conditioning.
\textbf{DriveTransformer}~\cite{jia2025drivetransformer}: a unified multi-task Transformer with parallel queries; input is multi-view camera with NC navigation conditioning.
\textbf{ORION}~\cite{fu2025orion}: a vision-language E2E policy for driving instruction generation; input is camera-only modality with NC navigation-command conditioning.
\textbf{DiffusionDrive}~\cite{liao2025diffusiondrive}: a truncated-diffusion planner for multimodal trajectory generation; the official implementation uses a ResNet backbone with real-time inference.
\textbf{SimLingo}~\cite{renz2025simlingo}: a language-guided E2E driving policy that aligns natural-language prompts with driving action sequences for interpretable human-in-the-loop control.
\textbf{KnowVal}~\cite{xia2025knowval}: a knowledge-augmented and value-guided driving system that retrieves open-world knowledge and evaluates candidate trajectories with a learned value model.
All methods are trained on the Bench2Drive training data.
For closed-loop evaluation, all methods are run on the 330 test routes described in Section~\ref{sec:method}, and metrics are computed accordingly. 

\subsection{Results}

\begin{table*}[t]
\caption{HiDrive full-eval results. We report split-level $DS$, $RC$, $LS$, and $ES$ for Overall and Basic/Hard/Thorny sets. All values are reported as 
percentages (\%).}
\label{tab:main_hlads}
\setlength{\tabcolsep}{3.5pt}
\renewcommand{\arraystretch}{1.05}
\centering\footnotesize
\begin{tabular}{@{}l|cccc|cccc|cccc|cccc@{}}
\toprule
\multirow{2}{*}{Method}
& \multicolumn{4}{c}{Overall}
& \multicolumn{4}{c}{Basic Set}
& \multicolumn{4}{c}{Hard Set}
& \multicolumn{4}{c}{Thorny Set} \\
\cmidrule(lr){2-5}\cmidrule(lr){6-9}\cmidrule(lr){10-13}\cmidrule(lr){14-17}
& $DS$ & $RC$ & $LS$ & $ES$
& $DS$ & $RC$ & $LS$ & $ES$
& $DS$ & $RC$ & $LS$ & $ES$
& $DS$ & $RC$ & $LS$ & $ES$ \\
\midrule
\rowcolor[gray]{0.93} TCP~\cite{wu2022trajectory}              
& 18.4 & 23.7 & 27.8 & 89.3 & 18.5 & 24.1 & 27.2 & 91.0 & 17.6 & 21.4 & 28.0 & 87.1 & 19.1 & 25.5 & 30.1 & 85.0 \\
UniAD-Base~\cite{UniAD}       
& 21.9 & 26.6 & 32.1 & 91.6 & 21.2 & 26.7 & 31.6 & 93.7 & 23.6 & 25.3 & 30.8 & 89.8 & 22.3 & 28.0 & 36.4 & 85.2 \\
\rowcolor[gray]{0.93} VAD~\cite{jiang2023vad}              
& 20.1 & 24.3 & 30.7 & 89.3 & 19.2 & 24.2 & 30.0 & 91.2 & 22.4 & 24.6 & 31.1 & 87.3 & 20.0 & 24.6 & 33.0 & 84.2 \\
ThinkTwice~\cite{jia2023think}       
& 28.3 & 35.6 & 38.9 & 92.4 & 26.3 & 36.8 & 38.3 & 95.2 & 33.5 & 31.8 & 39.3 & 88.7 & 29.0 & 36.0 & 41.3 & 86.0 \\
\rowcolor[gray]{0.93} DriveAdapter~\cite{jia2023driveadapter}    
& 29.5 & 40.0 & 40.3 & 92.4 & 27.6 & 41.8 & 40.0 & 95.3 & 34.0 & 34.7 & 39.0 & 89.0 & 30.8 & 40.3 & 43.7 & 84.5 \\
DriveTrans.~\cite{jia2025drivetransformer} 
& 31.0 & 40.3 & 43.1 & 91.9 & 28.9 & 40.2 & 42.7 & 94.5 & 36.6 & 36.2 & 43.1 & 88.5 & 31.3 & 47.0 & 45.0 & 85.8 \\
\rowcolor[gray]{0.93} ORION~\cite{fu2025orion}
& 37.4 & 65.3 & 54.8 & 93.5 & 34.4 & 62.9 & 51.4 & 95.5 & 43.8 & 70.1 & 61.3 & 91.4 & 40.5 & 68.7 & 59.5 & 87.8 \\
Diffu.Drive~\cite{liao2025diffusiondrive}   
& 34.2 & 58.3 & 49.1 & 91.4 & 30.8 & 56.0 & 45.3 & 93.4 & 41.3 & 64.0 & 56.4 & 89.4 & 38.1 & 59.6 & 54.8 & 85.3 \\
\rowcolor[gray]{0.93} SimLingo~\cite{renz2025simlingo}
& 42.3 & 70.5 & 63.3 & 94.8 & 39.1 & 66.7 & 60.9 & 96.3 & 50.4 & 77.4 & 70.7 & 92.1 & 43.8 & 76.2 & 62.1 & 92.7 \\
KnowVal~\cite{xia2025knowval}
& 46.6 & 73.8 & 69.3 & 97.4 & 42.6 & 70.5 & 66.0 & 98.3 & 56.8 & 79.9 & 78.5 & 96.3 & 47.9 & 79.0 & 69.4 & 95.3 \\
\bottomrule
\end{tabular}
\end{table*}


Table~\ref{tab:main_hlads} summarizes the overall performance of the completed runs on HiDrive, together with split-level results on the Basic, Hard, and Thorny ability sets.
The results show that conventional end-to-end autonomous driving methods remain strongly limited on HiDrive: even the best, DiffusionDrive, achieves only 34.2 Overall $DS$.
This indicates that, in a closed-loop environment involving long-tail objects, complex interactions, and legal and ethical constraints, methods that mainly rely on perception-prediction-planning pipelines or trajectory-generation paradigms still struggle to generalize to high-level driving capability evaluation.

In contrast, VLA and language-enhanced paradigms exhibit greater closed-loop robustness and generalization.
ORION and SimLingo both outperform the conventional end-to-end baselines, while KnowVal achieves the best overall performance with $DS$ = 46.6.
Compared with SimLingo, KnowVal further improves $DS$ by 4.3 points and $LS$ by 6.0 points.
These results suggest that injecting external or structured knowledge and using value-guided evaluation can help models make more stable decisions while adhering to legal compliance, risk avoidance, and ethical constraints.

One seemingly counterintuitive observation is that the $DS$ or $RC$ on the Hard and Thorny sets is often higher than that on the Basic set.
This does not imply that Hard or Thorny scenarios are inherently easier.
Instead, it is related to the ability taxonomy and metric definitions in HiDrive.
The Basic Set contains many tasks that are highly sensitive to low-level control stability and local interactions, in which collisions, deadlocks, or route interruptions can easily reduce $RC$ and $DS$.
By contrast, some Hard and Thorny scenarios emphasize high-level decision-making.
For example, when passing through a puddle, a vehicle may continue driving even if it does not fully comply with the intended ethical behavior, so the route is less likely to be interrupted.
This can lead to a higher $RC$ and consequently a higher $DS$.
However, Hard and Thorny scenarios are not easier: for most methods, $ES$ decreases from the Basic set to the Hard and Thorny sets, indicating that legal and ethical capabilities remain challenging.

Overall, HiDrive reveals clear differences among current autonomous driving paradigms: conventional end-to-end methods still show substantial limitations in long-tail closed-loop generalization, legal compliance, and social-norm understanding, while VLA-style methods benefit from stronger semantic understanding and instruction alignment.
Furthermore, the experimental results of KnowVal demonstrate that external knowledge injection and value-guided reasoning can further improve legal, ethical, and generalization capabilities.

\section{Conclusion}
\label{sec:conclusion}

In this paper, we propose HiDrive, a new closed-loop benchmark that systematically incorporates diverse rare objects and a richer set of driving capabilities, including rule compliance, moral reasoning, and emergency response. By extending evaluation beyond collision avoidance to encompass collision and braking behavior, traffic-rule compliance, and moral-reasoning indicators, HiDrive provides a more holistic assessment of driving intelligence. Our experiments reveal that state-of-the-art planners, which have already achieved near-saturated scores on widely used benchmarks, still exhibit significant performance degradation on HiDrive, particularly in long-tail perception and norm-sensitive scenarios. This highlights that current models, while highly competent in routine situations, are far from robust in the complex, ambiguous, and interactive conditions that characterize real-world driving. Built on a more advanced physics-rendering engine, HiDrive offers high-fidelity visual realism and physically plausible dynamics, making it a challenging and diagnostically valuable testbed for future research. We believe HiDrive will encourage the community to move beyond chasing incremental gains on saturated benchmarks and instead focus on building autonomous driving systems that are genuinely safe, socially aware, and prepared for the rare but critical events of everyday traffic.

\small
\bibliographystyle{plain}
\bibliography{references}

\begin{thebibliography}{10}

\bibitem{nuScenes}
Holger Caesar, Varun Bankiti, Alex~H. Lang, Sourabh Vora, Venice~Erin Liong, Qiang Xu, Anush Krishnan, Yu~Pan, Giancarlo Baldan, and Oscar Beijbom.
\newblock {nuScenes}: A multimodal dataset for autonomous driving.
\newblock In {\em CVPR}, 2020.

\bibitem{cao2025pseudo}
Wei Cao, Marcel Hallgarten, Tianyu Li, Daniel Dauner, Xunjiang Gu, Caojun Wang, Yakov Miron, Marco Aiello, Hongyang Li, Igor Gilitschenski, Boris Ivanovic, Marco Pavone, Andreas Geiger, and Kashyap Chitta.
\newblock Pseudo-simulation for autonomous driving.
\newblock In {\em CoRL}, 2025.

\bibitem{chen2026devil}
Canyu Chen, Yuguang Yang, Zhewen Tan, Yizhi Wang, Ruiyi Zhan, Haiyan Liu, Xuanyao Mao, Jason Bao, Xinyue Tang, Linlin Yang, Bingchuan Sun, Yan Wang, and Baochang Zhang.
\newblock Devil is in narrow policy: Unleashing exploration in driving {VLA} models.
\newblock {\em arXiv preprint arXiv:2603.06049}, 2026.

\bibitem{navsim}
Daniel Dauner, Marcel Hallgarten, Tianyu Li, Xinshuo Weng, Zhiyu Huang, Zetong Yang, Hongyang Li, Igor Gilitschenski, Boris Ivanovic, Marco Pavone, Andreas Geiger, and Kashyap Chitta.
\newblock {NAVSIM}: Data-driven non-reactive autonomous vehicle simulation and benchmarking.
\newblock In {\em NeurIPS}, 2024.

\bibitem{dosovitskiy2017carla}
Alexey Dosovitskiy, German Ros, Felipe Codevilla, Antonio Lopez, and Vladlen Koltun.
\newblock {CARLA}: An open urban driving simulator.
\newblock In {\em CoRL}, 2017.

\bibitem{fu2025orion}
Haoyu Fu, Diankun Zhang, Zongchuang Zhao, Jianfeng Cui, Dingkang Liang, Chong Zhang, Dingyuan Zhang, Hongwei Xie, Bing Wang, and Xiang Bai.
\newblock {ORION}: A holistic end-to-end autonomous driving framework by vision-language instructed action generation.
\newblock In {\em ICCV}, 2025.

\bibitem{hou2025driveagent}
Xinmeng Hou, Wuqi Wang, Long Yang, Hao Lin, Jinglun Feng, Haigen Min, and Xiangmo Zhao.
\newblock {DriveAgent}: Multi-agent structured reasoning with {LLM} and multimodal sensor fusion for autonomous driving.
\newblock {\em IEEE RAL}, 2025.

\bibitem{hu2022st}
Shengchao Hu, Li~Chen, Penghao Wu, Hongyang Li, Junchi Yan, and Dacheng Tao.
\newblock {ST-P3}: End-to-end vision-based autonomous driving via spatial-temporal feature learning.
\newblock In {\em ECCV}, 2022.

\bibitem{UniAD}
Yihan Hu, Jiazhi Yang, Li~Chen, Keyu Li, Chonghao Sima, Xizhou Zhu, Siqi Chai, Senyao Du, Tianwei Lin, Wenhai Wang, Lewei Lu, Xiaosong Jia, Qiang Liu, Jifeng Dai, Yu~Qiao, and Hongyang Li.
\newblock Planning-oriented autonomous driving.
\newblock In {\em CVPR}, 2023.

\bibitem{jia2023driveadapter}
Xiaosong Jia, Yulu Gao, Li~Chen, Junchi Yan, Patrick~Langechuan Liu, and Hongyang Li.
\newblock {DriveAdapter}: Breaking the coupling barrier of perception and planning in end-to-end autonomous driving.
\newblock In {\em ICCV}, 2023.

\bibitem{jia2023think}
Xiaosong Jia, Penghao Wu, Li~Chen, Jiangwei Xie, Conghui He, Junchi Yan, and Hongyang Li.
\newblock Think twice before driving: Towards scalable decoders for end-to-end autonomous driving.
\newblock In {\em CVPR}, 2023.

\bibitem{jia2024bench2drive}
Xiaosong Jia, Zhenjie Yang, Qifeng Li, Zhiyuan Zhang, and Junchi Yan.
\newblock {Bench2Drive}: Towards multi-ability benchmarking of closed-loop end-to-end autonomous driving.
\newblock In {\em NeurIPS}, 2024.

\bibitem{jia2025drivetransformer}
Xiaosong Jia, Junqi You, Zhiyuan Zhang, and Junchi Yan.
\newblock {DriveTransformer}: Unified transformer for scalable end-to-end autonomous driving.
\newblock In {\em ICLR}, 2025.

\bibitem{jiang2023vad}
Bo~Jiang, Shaoyu Chen, Qing Xu, Bencheng Liao, Jiajie Chen, Helong Zhou, Qian Zhang, Wenyu Liu, Chang Huang, and Xinggang Wang.
\newblock {VAD}: Vectorized scene representation for efficient autonomous driving.
\newblock In {\em ICCV}, 2023.

\bibitem{BEVFormer}
Zhiqi Li, Wenhai Wang, Hongyang Li, Enze Xie, Chonghao Sima, Tong Lu, Yu~Qiao, and Jifeng Dai.
\newblock {BEVFormer}: Learning bird's-eye-view representation from multi-camera images via spatiotemporal transformers.
\newblock In {\em ECCV}, 2022.

\bibitem{liao2025diffusiondrive}
Bencheng Liao, Shaoyu Chen, Haoran Yin, Bo~Jiang, Cheng Wang, Sixu Yan, Xinbang Zhang, Xiangyu Li, Ying Zhang, Qian Zhang, and Xinggang Wang.
\newblock {DiffusionDrive}: Truncated diffusion model for end-to-end autonomous driving.
\newblock In {\em CVPR}, 2025.

\bibitem{nie2024reason2drive}
Ming Nie, Renyuan Peng, Chunwei Wang, Xinyue Cai, Jianhua Han, Hang Xu, and Li~Zhang.
\newblock {Reason2Drive}: Towards interpretable and chain-based reasoning for autonomous driving.
\newblock In {\em ECCV}, 2024.

\bibitem{renz2025simlingo}
Katrin Renz, Long Chen, Elahe Arani, and Oleg Sinavski.
\newblock {SimLingo}: Vision-only closed-loop autonomous driving with language-action alignment.
\newblock In {\em CVPR}, 2025.

\bibitem{sima2024drivelm}
Chonghao Sima, Katrin Renz, Kashyap Chitta, Li~Chen, Hanxue Zhang, Chengen Xie, Jens Bei{\ss}wenger, Ping Luo, Andreas Geiger, and Hongyang Li.
\newblock {DriveLM}: Driving with graph visual question answering.
\newblock In {\em ECCV}, 2024.

\bibitem{sun2026sparsedrivev2}
Wenchao Sun, Xuewu Lin, Keyu Chen, Zixiang Pei, Xiang Li, Yining Shi, and Sifa Zheng.
\newblock {SparseDriveV2}: Scoring is all you need for end-to-end autonomous driving.
\newblock {\em arXiv preprint arXiv:2603.29163}, 2026.

\bibitem{sun2025sparsedrive}
Wenchao Sun, Xuewu Lin, Yining Shi, Chuang Zhang, Haoran Wu, and Sifa Zheng.
\newblock {SparseDrive}: End-to-end autonomous driving via sparse scene representation.
\newblock In {\em ICRA}, 2025.

\bibitem{wang2026latent}
Linbo Wang, Yupeng Zheng, Qiang Chen, Shiwei Li, Yichen Zhang, Zebin Xing, Qichao Zhang, Xiang Li, Deheng Qian, Pengxuan Yang, Yihang Dong, Ce~Hao, Xiaoqing Ye, Junyu Han, Yifeng Pan, and Dongbin Zhao.
\newblock {Latent-WAM}: Latent world action modeling for end-to-end autonomous driving.
\newblock {\em arXiv preprint arXiv:2603.24581}, 2026.

\bibitem{weng2024drive}
Xinshuo Weng, Boris Ivanovic, Yan Wang, Yue Wang, and Marco Pavone.
\newblock {PARA-Drive}: Parallelized architecture for real-time autonomous driving.
\newblock In {\em CVPR}, 2024.

\bibitem{wu2022trajectory}
Penghao Wu, Xiaosong Jia, Li~Chen, Junchi Yan, Hongyang Li, and Yu~Qiao.
\newblock Trajectory-guided control prediction for end-to-end autonomous driving: A simple yet strong baseline.
\newblock In {\em NeurIPS}, 2022.

\bibitem{xia2025knowval}
Zhongyu Xia, Wenhao Chen, Yongtao Wang, and Ming-Hsuan Yang.
\newblock {KnowVal}: A knowledge-augmented and value-guided autonomous driving system.
\newblock {\em arXiv preprint arXiv:2512.20299}, 2025.

\bibitem{henetpp}
Zhongyu Xia, Zhiwei Lin, Yongtao Wang, and Ming-Hsuan Yang.
\newblock {HENet++}: Hybrid encoding and multi-task learning for 3d perception and end-to-end autonomous driving.
\newblock {\em arXiv preprint arXiv:2511.07106}, 2025.

\bibitem{xu2024drivegpt4}
Zhenhua Xu, Yujia Zhang, Enze Xie, Zhen Zhao, Yong Guo, Kwan-Yee~K. Wong, Zhenguo Li, and Hengshuang Zhao.
\newblock {DriveGPT4}: Interpretable end-to-end autonomous driving via large language model.
\newblock {\em IEEE RAL}, 2024.

\bibitem{yao2025comal}
Huaiyuan Yao, Longchao Da, Vishnu Nandam, Justin Turnau, Zhiwei Liu, Linsey Pang, and Hua Wei.
\newblock {CoMAL}: Collaborative multi-agent large language models for mixed-autonomy traffic.
\newblock In {\em SDM}, 2025.

\bibitem{zheng2024genad}
Wenzhao Zheng, Ruiqi Song, Xianda Guo, Chenming Zhang, and Long Chen.
\newblock {GenAD}: Generative end-to-end autonomous driving.
\newblock In {\em ECCV}, 2024.

\end{thebibliography}



\end{document}